\pdfoutput=1
\PassOptionsToPackage{prologue,dvipsnames}{xcolor}
\documentclass[11pt]{article}

\usepackage[preprint]{acl}

\usepackage{times}
\usepackage{latexsym}
\usepackage{makecell}
\usepackage{booktabs}
\usepackage[T1]{fontenc}

\usepackage[utf8]{inputenc}

\usepackage{microtype}

\usepackage{inconsolata}

\usepackage{graphicx}

\usepackage{adjustbox}
\usepackage{multirow}
\usepackage{multicol}
\usepackage{booktabs}
\usepackage{setspace}
\usepackage{hyperref}
\usepackage{threeparttable}
\usepackage{colortbl}
\usepackage{float}
\usepackage{enumitem}
\usepackage{array}
\usepackage{arydshln}
\usepackage{makecell}
\usepackage{amsmath}
\usepackage{subfigure}
\newcommand{\method}{{MetaLadder}}

\usepackage{algorithm}
\usepackage{algpseudocode}
\usepackage{colortbl}

\usepackage{amssymb}
\usepackage{mathtools}
\usepackage{amsthm}

\usepackage{colortbl}
\usepackage{nicematrix}

\usepackage{amsthm}
\usepackage{amsmath,amsfonts,bm}
\usepackage{nicefrac}
\usepackage{array}

\usepackage{url}
\usepackage{pbox}

\definecolor{bluex}{rgb}{0.27, 0.42, 0.81}
\definecolor{purplex}{HTML}{9564bf}
\definecolor{red3}{HTML}{C52A20}
\definecolor{red2}{HTML}{B36A6F}
\definecolor{red1}{HTML}{FFb5b5}
\definecolor{purple}{HTML}{B36A6F}
\definecolor{darkyellow}{HTML}{D5BA82}
\definecolor{blue1}{HTML}{508AB2}
\definecolor{blue2}{HTML}{C4E4E3}
\definecolor{green1}{HTML}{A1D0C7}
\definecolor{green2}{HTML}{BFF6BA}
\definecolor{green3}{HTML}{028100}
\definecolor{teal}{HTML}{508AB2}
\definecolor{purple1}{HTML}{8d3a94}

\usepackage{pifont}

\usepackage[listings]{tcolorbox}
\tcbuselibrary{listings,theorems}
\newtcolorbox{mybox}{colback=white!5!white,colframe=black!75!black, left=.05in, right=.05in}

\newtcbtheorem[number within=section]{exmp}{Example}%
{beforeafter skip balanced=.01in,colback=green2!5,colframe=blue1,fonttitle=\small \bfseries, left=.02in, right=.02in,bottom=.02in, top=.02in}{exmp}

\theoremstyle{plain}

\theoremstyle{definition}

\theoremstyle{remark}

\title{MetaLadder: Ascending Mathematical Solution Quality via\\ Analogical-Problem Reasoning Transfer}

\author{
    Honglin Lin\textsuperscript{1},
    {\bf Zhuoshi Pan\textsuperscript{1,2}},
    {\bf Yu Li\textsuperscript{1}},
    {\bf Qizhi Pei\textsuperscript{1,3}},
    {\bf Xin Gao\textsuperscript{1}},\\
    {\bf Mengzhang Cai\textsuperscript{1}},
    {\bf Conghui He\textsuperscript{1}},
    {\bf Lijun Wu\textsuperscript{1}}\thanks{\ \ Corresponding author.} \\
    \textsuperscript{1}Shanghai AI Laboratory \quad \textsuperscript{2}Tsinghua University \quad
    \textsuperscript{3}Renmin University of China \\
    \texttt{\{linhonglin,wulijun\}@pjlab.org.cn} 
}

\begin{document}
\maketitle
\begin{abstract}

Large Language Models (LLMs) have demonstrated promising capabilities in solving mathematical reasoning tasks, leveraging Chain-of-Thought (CoT) data as a vital component in guiding answer generation. Current paradigms typically generate CoT and answers directly for a given problem, diverging from human problem-solving strategies to some extent. Humans often solve problems by recalling analogous cases and leveraging their solutions to reason about the current task. Inspired by this cognitive process, we propose \textbf{MetaLadder}, a novel framework that explicitly prompts LLMs to recall and reflect on meta-problems, those structurally or semantically analogous problems, alongside their CoT solutions before addressing the target problem. Additionally, we introduce a problem-restating mechanism to enhance the model’s comprehension of the target problem by regenerating the original question, which further improves reasoning accuracy. Therefore, the model can achieve reasoning transfer from analogical problems, mimicking human-like “learning from examples” and generalization abilities. Extensive experiments on mathematical benchmarks demonstrate that our MetaLadder significantly boosts LLMs’ problem-solving accuracy, largely outperforming standard CoT-based methods (\textbf{10.3\%} accuracy gain) and other methods. Our code and data has been released at \url{https://github.com/LHL3341/MetaLadder}.

\end{abstract}
\begin{figure}[!ht]
    \centering
    \includegraphics[width=0.75\linewidth]{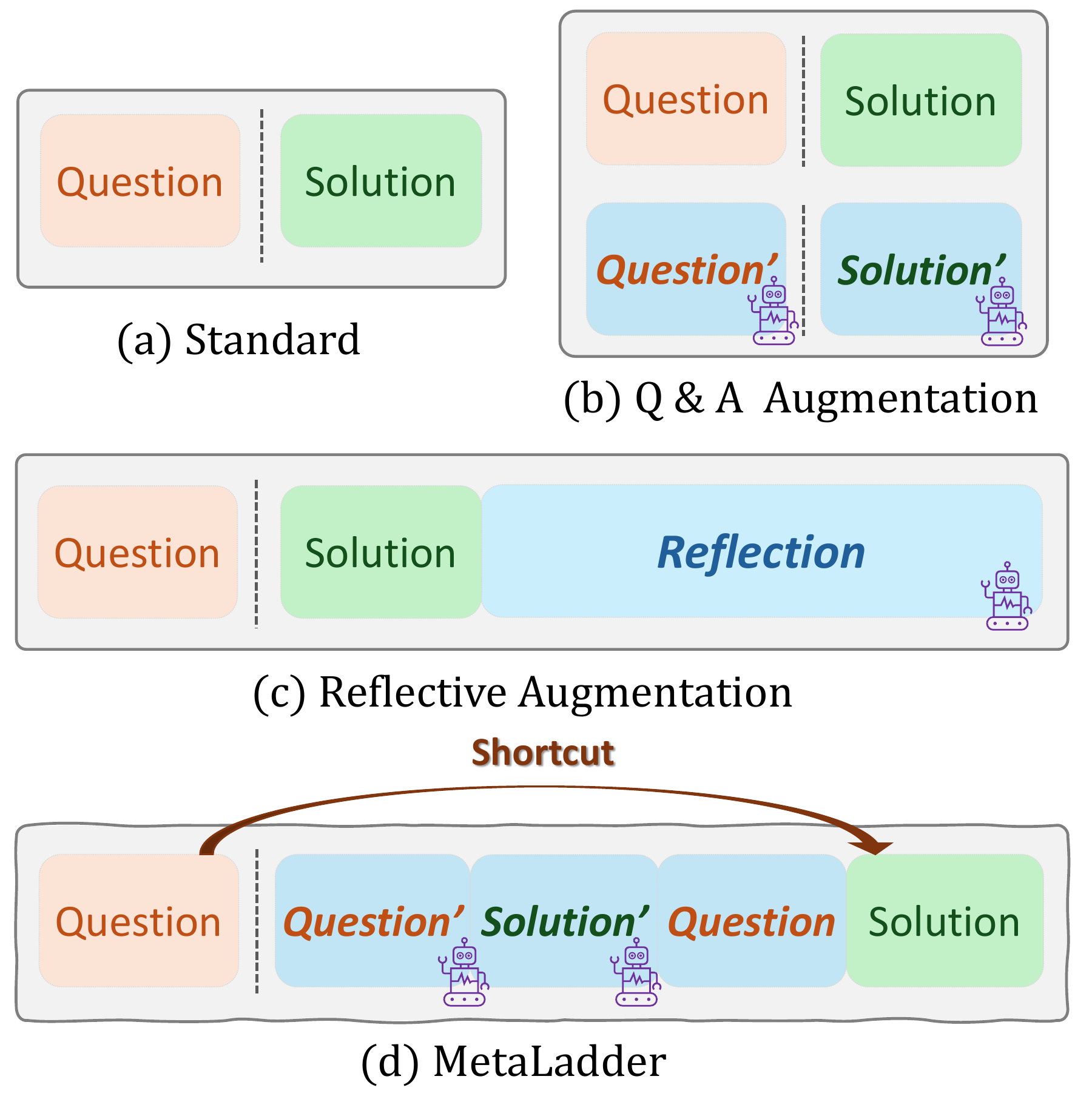}
    \caption{Compare our MetaLadder with other methods (Standard CoT, Question \& Answer Augmentation, Reflective Augmentation) on training data construction.}
    \label{fig:compare}
\end{figure}

\section{Introduction}

Large Language Models (LLMs) have achieved remarkable success in mathematical reasoning tasks by leveraging Chain-of-Thought (CoT) data, which explicitly guides models to decompose problems into intermediate reasoning steps before producing final answers~\cite{openai_o1,deepseek_r1,qwq-32b-preview}. Pioneering works such as ~\cite{cot_reasoning} demonstrated that training LLMs on CoT-style solutions significantly improves their ability to solve complex problems, with subsequent studies~\cite{fu2022complexity,zhou2022least} further refining this paradigm. For instance, models like Minerva and GPT-4~\cite{lewkowycz2022solving,openai2023gpt4} have showcased near-human performance by distilling high-quality CoT trajectories from expert demonstrations. These methods typically follow a straightforward template: given a problem, the model generates a CoT explanation step-by-step, which then leads to the correct answer. While effective, such approaches align only partially with the nuanced cognitive processes~\cite{thinking} humans employ during problem-solving.

Despite their success, existing CoT-based fine-tuning methods rely on a rigid ``Problem $\to$ CoT $\to$ Answer'' framework~\cite{small-lm-math-reason, metamath}, which diverges from how humans approach challenging mathematical tasks. When solving problems, humans rarely generate solutions in isolation; instead, they actively recall analogous problems and their solutions, especially for difficult or unfamiliar questions~\cite{vosniadou1988analogical, daugherty2008analogical}. 
For example, encountering a combinatorics problem, a student might recall similar problems involving permutations or recursive strategies, using their structures to guide the current solution. This ability to leverage prior analogical experiences is critical for generalizing knowledge and tackling novel challenges. However, current LLM training paradigms largely overlook this aspect, treating each problem as an independent instance without encouraging cross-problem reasoning. This limitation constrains models' capacity to transfer learned reasoning patterns, particularly for problems requiring abstract structural or semantic similarities to prior examples.

To bridge this gap, we propose \textbf{MetaLadder}, a framework inspired by human-like analogical reasoning and problem comprehension. MetaLadder explicitly guides LLMs to \textit{recall and reflect on meta-problems}—structurally or semantically analogous problems with known CoT solutions—before generating answers for the target problem. 
These meta-problems and their CoT trajectories serve as scaffolding to derive the current solution, mirroring how humans ``stand on the shoulders'' of past experiences. Additionally, we introduce a \textit{problem-restating mechanism}: before reasoning, the model regenerates the original question in its own words, enhancing its comprehension of the problem's core components and constraints. This dual mechanism—analogical recall and active restatement—enables the model to decompose complex problems into familiar reasoning patterns, effectively mimicking the human ability to ``learn from examples'' and generalize solutions across analogous contexts. By integrating these steps, MetaLadder successfully transforms the traditional linear CoT process into a dynamic, context-aware reasoning ladder, where each rung represents a retrieved meta-problem or a refined understanding of the target task.

Extensive experiments validate MetaLadder's effectiveness. On mathematical benchmarks like GSM8K and MATH, models trained with MetaLadder achieve significant improvements over standard CoT fine-tuning baselines, with accuracy gains of 12.4\% and 11.3\%, respectively, surpassing recent advanced methods. 
Further analysis reveals that MetaLadder-trained models exhibit stronger generalization to structurally novel problems, solving 9.3\% more ``out-of-distribution'' test cases than vanilla CoT models. 
Qualitative examples demonstrate that the model not only retrieves relevant meta-problems but also adapts their solutions creatively.
These results collectively highlight that emulating human-like analogical reasoning and active comprehension is a powerful yet underexplored direction for advancing LLMs' mathematical reasoning capabilities.

\section{Related Work}
\subsection{Data Synthesis for Math Reasoning} Data synthesis has significantly contributed to the development of LLMs' mathematical reasoning abilities~\cite{qwen25_math, deepseekmath}. Some studies focus on expanding the dataset and its diversity by rewriting questions or answers~\cite{metamath,yuan2023rft,liu2024mmiqc,tang2024mathscale, wizardmath}. For example, MetaMath~\cite{metamath} diversifies the data through various enhancement methods, including question rephrasing, answer augmentation, and the generation of inverse problems.
Another line of research focuses on improving the quality and difficulty of the data~\cite{wang2024mathcoder, wizardmath, dartmath,refaug, reformat-alignment}. For instance, WizardMath~\cite{wizardmath} generates more challenging data through RLEIF, while RefAug~\cite{refaug} adds reflective information after the original CoT process to encourage enhancing the reasoning process.
Our method differs by augmenting data to activate the model's analogical reasoning capabilities, enabling the model to generate and apply solutions based on analogous problems rather than relying solely on paraphrased data, even enabling self-evolution by generating analogous data.

\begin{figure*}[t]
    \centering
    \includegraphics[width=0.9\linewidth]{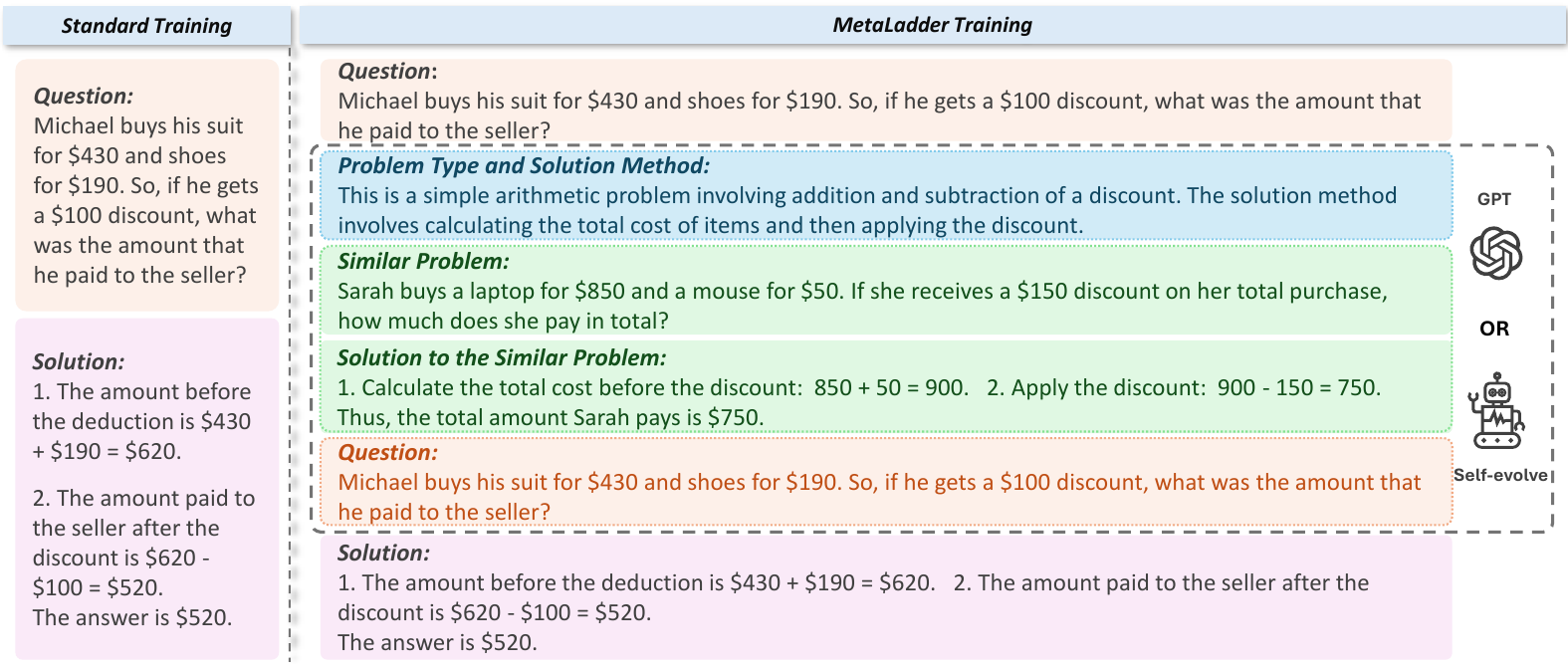}
    \caption{Overview of the MetaLadder framework for generating reflective data. The process starts with the original problem $Q$, followed by the problem type and solution method $S$, and the generation of analogous problems $Q'$ and solutions $C'$. Afterward, the original problem $Q$ is reintroduced to prompt the model to restate the problem. These components are then combined with the solution $C$ of the original problem $Q$ to form the training data.}
    \label{fig:method}
\end{figure*}

\subsection{RAG for Problem Solving}
Retrieval-Augmented Generation (RAG) systems enhance the performance of LLMs by integrating an external search engine for knowledge retrieval~\cite{khandelwal2019generalization, rag, rag_survey}. When a user poses a question, the RAG system first retrieves relevant knowledge fragments through the search engine and then uses these answers along with the original query to generate an answer.
To address more complex problems, such as mathematical reasoning, some works incorporate the reasoning capabilities of LLMs into the RAG framework, achieving retrieval-augmented reasoning (RAR)~\cite{melz2023enhancingllmintelligencearmrag}. For example, IRCoT~\cite{ircot} combines RAG with multi-step CoT by using the question and previous reasoning steps as queries, retrieving relevant documents to generate the next reasoning step. RAT~\cite{rat} generates a complete CoT first and then refines each reasoning step iteratively using RAG.
Recent work, such as Search-o1~\cite{search-o1}, extends the RAG paradigm by applying it to o1-like models, further enhancing the model's reasoning capabilities.
While RAG emphasizes enhancing the model's performance by retrieving external knowledge without updating the model's parameters, our approach differs in that it internalizes analogical reasoning through model fine-tuning, allowing the model to generate reflective information during reasoning without relying on external data.

\section{\method}

We first introduce the overall MetaLadder framework for enhancing mathematical problem-solving (Section~\ref{method:overview}). Then we present each our method in detail by explaining the generation of reflective data to guide the model's reasoning process (Section~\ref{method:gen}), describing the composition of training data to activate the model's analogical reasoning (Section~\ref{method:train}). Besides, we also add a self-evolve process in our framework to enable the model's ability to autonomously generate data for self-improvement (Section~\ref{method:self-evolve}). Finally, we also incorporate a shortcut inference mechanism for fast and effective generation (Section~\ref{method:shortcut}). 

\subsection{Overview}
\label{method:overview}
The overview of MetaLadder framework is illustrated in Fig.~\ref{fig:method}. Given the original data consisting of a target problem $Q$ and its solution $C$. We first generate additional reflective data by synthesizing an problem analysis and solution strategy $S$ for the target problem, then along with analogous problem $Q'$ and its corresponding solution $C'$, which are structurally or semantically similar to the original data. Moreover, we introduce a problem-restating mechanism, where the target problem $Q$ is inserted before the final solution $C$ to enhance the model's understanding of the target problem. After training on the generated data sequence $QSQ'C'QC$, the model is able to recall analogous problems, restate the target problem and apply analogical reasoning to find the final solution.
Notably, because of MetaLadder's analogical reasoning capability, the model can autonomously generate similar problems for training itself, which enables the self-evolution ability of the model. We now introduce the details in the following sections.

\subsection{Reflective Data Generation}
\label{method:gen}
MetaLadder improves the model's mathematical problem-solving by incorporating reflective data during training. This approach simulates human learning, encouraging the model to recall and reflect on analogous problems, using their solutions to inform reasoning on the target problem. 
To achieve this, the model requires structured guidance to first understand the problem-solving strategy, then recall analogous problems, and finally apply the solutions from those analogous problems to the current task. Therefore, the reflective data consists of three key components:

\noindent \textbf{1) Problem Type and Solution Method $S$}. Each problem is categorized into a mathematical domain, with an explanation of the relevant concepts and methods (e.g., \textit{``This is a simple arithmetic problem involving addition and subtraction of a discount. The solution method involves calculating the total cost of items and then applying the discount.''}). This helps the model grasp the problem-solving framework for future similar problems.

\noindent \textbf{2) Analogous Problem $Q'$}. The model generates an analogous problem by modifying the context, numbers, or variables, while keeping the core structure intact, offering a new learning context. For example, as shown in Fig~\ref{fig:method}:
\textit{``Original Problem: Michael buys his suit for \$430 and shoes for \$190. So, if he gets a \$100 discount, what was the amount that he paid to the seller?''}
The generated analogy problem is then:
\textit{``Similar Problem: Sarah buys a laptop for \$850 and a mouse for \$50. If she receives a \$150 discount on her total purchase, how much does she pay in total?''}

\noindent \textbf{3) Solution to Analogous Problem $C'$}. The model provides a solution for the analogous problem, reinforcing the application of similar strategies across different problems.
For instance, the solution to the above analogy problem is:
\textit{``1. Calculate the total cost before the discount:  850 + 50 = 900. 2. Apply the discount:  900 - 150 = 750.  Thus, the total amount Sarah pays is \$750.''}

To generate the above annotation data, we carefully design prompts for data generation. The detailed prompts are provided in Table~\ref{annotation_prompt}.

\subsection{Analogical Reasoning Activation}
\label{method:train}
To activate the analogical reasoning capabilities of the model, we compose the training data with the generated reflective data in a format as described in Figure~\ref{fig:compare} (d). 
In traditional approaches, as shown in Fig.~\ref{fig:compare}, CoT directly generates the solution $C$ from the problem $Q$, while question and answer augmentation directly rephrase the problem and solution into $Q'C'$. RefAug~\cite{refaug} adds additional reflection $R$ after the solution $C$. In contrast, our MetaLadder introduces an analogical reasoning process $SQ'C'Q$ between $Q$ and $C$, involving the generation of analogous problem and the transfer of knowledge from similar problem to the target problem. Specifically, our analogical reasoning process $SQ'C'Q$ consists of:

\noindent\textbf{Problem Type and Solution Method} $S$. This is the problem and solution analysis, which is generated in Section~\ref{method:gen}.

\noindent\textbf{Analogical Problem and Corresponding Solution} $Q'C'$. This is the analogy problem and the solution of the analogy problem, which is also generated in Section~\ref{method:gen}.

\noindent\textbf{Problem-restating mechanism.} 
After presenting $Q'C'$, we reintroduce the original problem $Q$. This step ensures that the model revisits the target problem and apply the knowledge gained from solving the analogous problem, engaging in analogical reasoning and transferring the learned solution to solve the original problem.

Overall, the enhanced training data format is $QSQ'C'QC$: Original Problem $\to$ Problem Type and Solution Strategy $\to$ Analogous Problem and its Solution $\to$ Original Problem and its Solution. 
By training on the above formatted data, we aim to improve the model's mathematical reasoning by activating its analogical reasoning from similar problem and solution with a deeper understanding of the target problem.

\subsection{Analogical Self-evolution}
\label{method:self-evolve}
Similar to other works~\cite{star, luong2024reft, rstar_math}, after training, our model gains the ability to autonomously generate analogous problems that are related to the target problem. This capability facilitates a self-evolving data augmentation process, where the model can iteratively bootstrap its own dataset. Specifically, after making predictions for a given problem, the model is used to generate new problem instances based on its own outputs. These generated problems, being structurally or conceptually similar to the original ones, are then fed back into the training loop for further refinement.
The self-evolution process significantly enhances the model’s ability to expand its knowledge base. As the model generates new problem instances, it creates novel variations of existing problems by modifying key components—such as numbers, variables, or contexts—while preserving the underlying structure. This process not only reinforces the model's understanding of the problem-solving strategies but also improves its generalization ability across new, previously unseen problem types.

\subsection{Shortcut Inference}
\label{method:shortcut}

During training, the model learns to implicitly encode analogy-problem reasoning schemas via the $QSQ’C’QC$ paradigm, where explicit generation of analogical problems $Q’$ and their solutions $C’$ establishes robust neural pathways for structural pattern transfer. At inference time, we try to propose a shortcut inference method that enables a streamlined $QSQC$ process that bypasses $Q’C’$ generation for fast inference. Specifically, after the model generates $S$, we force it to directly restate the original problem $Q$ and output the answer $C$ by inserting $Q$ after $S$.
Surprisingly, skipping the $Q'C'$ generation not only reduces inference cost but also boost the performance significantly (see results in Section~\ref{sec:main}). This clearly demonstrates MetaLadder can successfully learn analogy knowledge transfer through analogy problem solving.

\section{Experiments}

\subsection{Experimental Setup}
\noindent\textbf{Datasets/Benchmarks}.
We use the training sets from GSM8k~\cite{cobbe2021gsm8k} and MATH~\cite{hendrycks2021math} for our experiments. 
The augmented parts in each problem (problem type and solution method, the analogy problem, the solution to the analogy problem) are generated by GPT-4o-mini~\cite{openai_4omini}. The details of the data generation process can be found in the Appendix~\ref{apx:gen_detail}.
For evaluation, besides the test sets from GSM8K and MATH as in-distribution evaluation, we also include out-of-distribution test sets from ASDiv~\cite{miao2020diverse}, College Math~\cite{tang2024mathscale}, GaoKao (Chinese College Entrance Exam) En 2023~\cite{liao2024mario} and DM~\cite{mathematic} for verification.

\noindent\textbf{Training and Evaluation}.
We primarily use two popular LLMs for our experiments, covering the general focused LLM and the math-focused LLM: the widely used Llama3-8B~\cite{llama3} and DeepSeekMath-7B~\cite{deepseekmath}. 
For a fair comparison, all the models are trained for 1 epoch. During inference, greedy decoding is applied to get the outputs. 
As for evaluation metrics, we report Pass@1 accuracy for all the models and baselines. 
More experimental details can be found in the Appendix~\ref{apx:train_detail} and \ref{apx:evaluation_detail}.

\noindent{\textbf{Baselines}}.
We first introduce the following baseline methods that we adopt for comparison, which are also shown in Figure~\ref{fig:compare}:
\textbf{(i) CoT:} The original CoT data from GSM8K and MATH, which is the standard setting (Figure~\ref{fig:compare}(a)).
\textbf{(ii) AnalogyAug:} This combines our augmented analogy problem/solution (as used in MetaLadder) with the original CoT data in a \textit{batch-level} training, making the training data twice as large as the original data (also known as question\&answer augmentation, shown in Figure~\ref{fig:compare}(b)).
\textbf{(iii) RefAug:} A state-of-the-art (SOTA) data augmentation method that enhances model reasoning by \textit{appending} reflective data to the end of the CoT chain (Figure~\ref{fig:compare}(c)).

For our MetaLadder-based settings, we enhance the original CoT data as described in Section~\ref{method:train} to train \textbf{1) MetaLadder}, the basic setting, and derive the following variant
\textbf{2) MetaLadder + Self-evolve}: We use the MetaLadder model after one round of training to greedily sample one data point from each problem, and then filter out the correct answers to add back to the training data used in the first round. 
Since we have generated the analogy problem $Q'$ for helping solve the target problem $Q$, we add a reverse training setting in this section. Specifically, we train \textbf{3) MetaLadder + Reverse}, which simply swaps the target problem $Q$ and the analogous problem $Q'$ to be a new training sample, expanding the training data to twice the size of the original data. Besides, we also experiment on a variant, \textbf{4) MetaLadder + Reverse + Self-evolve}, which further incorporates self-evolve data.
Besides, for our MetaLadder-related experiments, we also include the shortcut inference mechanism as a comparison. The "-cut" suffix after the method name indicates the use of the shortcut inference.

\subsection{Main Results}
\label{sec:main}

\begin{table*}[]
\centering
\renewcommand{\arraystretch}{0.9}
\resizebox{0.9\textwidth}{!}{
\begin{tabular}{lcccccccc}
\toprule
\multicolumn{2}{c}{}  & \multicolumn{2}{c}{\textbf{In-Domain}} & \multicolumn{4}{c}{\textbf{Out-of-Domain}}  \\ \cmidrule(rl){3-4} \cmidrule(rl){5-8} 
\multicolumn{1}{c}{\multirow{-2}{*}{\textbf{Method}}} & \multicolumn{1}{c}{\multirow{-2}{*}{\textbf{\# Sample}}} & GSM8K     & MATH     & ASDiv & CM & GE & DM & Average\\ \hline
\multicolumn{9}{c}{\cellcolor[HTML]{C0C0C0}\textbf{\textit{LLaMA3-8B}}}       \\
\hline
CoT~\cite{cot_reasoning}  &15K  & 61.5     & 19.4    & 73.2  & 16.6 & 19.5 & 26.3 & 36.1\\
RefAug~\cite{refaug}      &15K  & 59.7     & 20.3    & 74.3  & 17.6 & 18.4 & 23.2 & 35.6\\
RefAug+CoT~\cite{refaug}  &30K  &64.9      &21.8     &74.4   & 16.5 & 21.0 &24.8 &37.4\\
AnalogyAug                &30K  & 63.8     & 22.6    & 76.1  & 18.2 & 20.8 & 27.2 &38.1\\
\hdashline
MetaLadder                &15K  & 66.2     & 22.4    & 76.7  & 17.2 & 23.9 & 29.7 &39.4\\
\rowcolor{gray!20}
MetaLadder-cut           &15K  & 69.4     & 24.0    & \underline{78.8}  & 18.2 & 26.0 & 29.4 &41.0\\
MetaLadder+Self-evolve    &22K  & 66.5     & 24.8    & 76.6  & 19.1 &26.8 & \underline{31.7} &40.9\\
\rowcolor{gray!20}
MetaLadder+Self-evolve-cut &22K  & \textbf{73.5}   & 26.0    & \textbf{81.6}  & \textbf{20.3}  & 24.4 & 30.7 & \textbf{42.8} \\
MetaLadder+Reverse        &30K   &\underline{71.5}    &25.6 &77.2 & 19.0 & 24.7 & 29.0 &41.2\\
MetaLadder+Reverse+Self-evolve   &44K & 71.1 & \underline{26.7}    & 77.2  & 19.2 & \underline{27.3} & 31.4 & 42.2\\
\rowcolor{gray!20}
MetaLadder+Reverse+Self-evolve-cut   &44K & 70.5 & \textbf{27.7} & 77.2  & \underline{19.4} & \textbf{28.3} & \textbf{31.9} & \underline{42.5}\\

\hline
\multicolumn{9}{c}{\cellcolor[HTML]{C0C0C0}\textbf{\textit{DeepSeekMath-7B}}} \\
\hline
CoT~\cite{cot_reasoning}       &15K  & 64.2     & 34.3    & 81.5  & 31.4 & 29.4 & 43.0 & 47.3\\
RefAug~\cite{refaug}           &15K  & 67.4     & 35.1    & 83.0  & 30.3 & 35.6 & 43.5 & 49.2\\
RefAug+CoT~\cite{refaug}       &30K  & 67.2     & 34.9    & 80.4  & 31.8 & 29.4 & 45.1 & 48.1\\
AnalogyAug                     &30K  & 67.7     & 38.9    & 83.2  & 30.5 & 36.9 & 49.6 & 51.1\\
\hdashline
MetaLadder                     &15K  & 69.4     & 38.6    & 85.9  & 32.6 & 37.4 & 48.4 & 52.1\\
\rowcolor{gray!20}
MetaLadder-cut                 &15K  & 71.5     & 40.0    & 87.1  & \underline{35.2} & \underline{40.5} & 49.1 & 53.9\\
MetaLadder+Self-evolve         &23K  & 70.5     & 39.3    & 86.3  & 33.3 & 35.1 & 50.8 & 52.6\\
\rowcolor{gray!20}
MetaLadder+Self-evolve-cut    &23K  & \underline{74.1}  &  \underline{41.3}  & \underline{87.3}  & \textbf{35.7} & \underline{40.5} & 50.5 & \underline{54.9} \\
MetaLadder+Reverse             &30K  & 72.3  & 40.5 & 85.2  & 32.1 & 37.4 & 51.3 & 53.1\\
MetaLadder+Reverse+Self-evolve   &46K & 72.6 & 40.7 & 85.2  & 33.7 & 38.2 & \underline{51.9} & 53.7\\ 
\rowcolor{gray!20}
MetaLadder+Reverse+Self-evolve-cut & 46K & \textbf{76.6} & \textbf{45.6} & \textbf{89.3} & 35.1 & \textbf{43.1} & \textbf{54.8} & \textbf{57.6} \\
\bottomrule
\end{tabular}
}
\caption{Accuracy on in-domain and out-of-domain mathematical benchmarks. The \textbf{bold} and \underline{underlined} values represent the first and second best performances, respectively. CM, GE, DM denotes College Math, Gaokao En 2023, DeepMind-Mathematics, respectively.}
\label{exp:main}
\end{table*}

Our experimental results, as shown in Table~\ref{exp:main}, reveal the following key findings:

\noindent\textbf{MetaLadder Outperforms Strong Methods.} The main experimental results demonstrate the effectiveness of the MetaLadder framework across multiple mathematical benchmarks. MetaLadder consistently outperforms baseline methods on both in-domain and out-of-domain datasets. On the LLaMA3-8B and DeepSeekMath-7B models, MetaLadder surpasses CoT by an average accuracy improvement of 6.7 (36.1 vs. 42.8) and 10.3 (47.3 vs. 57.6) points, respectively, and outperforms RefAug by 5.4 and 9.5 points in accuracy. These results highlight MetaLadder's significant advantage in enhancing the model's reasoning ability, particularly when tackling challenging mathematical problems.

\noindent\textbf{MetaLadder Enhances the Model Beyond Batch-Level Augmentation.} Compared to AnalogyAug, which performs question and answer augmentation at the batch level, MetaLadder achieves higher scores on GSM8K and comparable performances on MATH. On both models, MetaLadder improves by 4.7 and 6.5 points, respectively. 
This suggests that MetaLadder effectively enhances the model's reasoning ability by activating its analogical reasoning capabilities, rather than simply adding more augmented data.
Particularly, MetaLadder + Reverse, which swaps the target problem with analogous problems to double the dataset, outperforms AnalogyAug by 8.1 and 4.6 points on the two models, respectively. This further validates the effectiveness of MetaLadder's data generation strategy.

\noindent\textbf{MetaLadder’s Self-Evolution Boosts Model Performance.} Self-evolution provides further improvements in model performance, with significant gains observed across all datasets. 
After one round of self-evolution, MetaLadder+Self-evolve improves by 1.5 points on LLaMA3-8B and 0.5 points on DeepSeekMath-7B, demonstrating that a single round of self-training effectively enhances the model's reasoning ability. 
Additionally, MetaLadder+Reverse+Self-evolve improves by 1.0 points and 0.6 points on LLaMA3-8B and DeepSeekMath-7B in accuracy across all datasets, respectively, confirming the benefits of data augmentation through problem swapping. Ultimately, MetaLadder + Reverse + Self-evolve exceeds CoT by 6.1 points and RefAug by 4.8 points on LLaMA3-8B and achieves the best score of 53.7 on DeepSeekMath-7B.

\noindent\textbf{Shortcut Inference Reduces Inference Cost and Improves Performance.} Surprisingly, shortcut inference not only reduces the inference cost by skipping the analogy problem reasoning during inference (e.g., as shown in Table~\ref{exp:ablation}, DeepSeek-MetaLadder-cut on MATH achieves 1343.74 seconds, faster than DeepSeek-MetaLadder's 2181.13 seconds and close to CoT's 1253.26 seconds), but also boosts the model performance by a clear margin, e.g., 1.6 accuracy points on LLaMA3-8B and 1.8 points on DeepSeekMath-7B. The results demonstrate MetaLadder has transferred analogy problem-solving knowledge.

These results underscore MetaLadder's outstanding performance in solving both in-domain and out-of-domain problems, further validating the framework's effectiveness in enhancing mathematical problem-solving abilities and cross-domain generalization. Through its data augmentation and self-evolution strategies, MetaLadder not only excels in reasoning tasks within known domains but also demonstrates strong adaptability when facing unfamiliar data.

\subsection{Ablation on Components}

To thoroughly evaluate the contribution of each component in the MetaLadder framework, we conducted an ablation study that systematically examined the impact of its core elements, where ``w/o Strategy'', ``w/o Analogy'', and ``w/o Restate'' refer to the absence of the problem type and solution method, analogy meta-problem, and problem restating mechanism, respectively.

\begin{table}[h]
    \centering
    \resizebox{0.4\textwidth}{!}{
    \begin{tabular}{lrrr}
        \toprule
        \textbf{Method} & \textbf{GSM8K} & \textbf{MATH} & \textbf{Average}\\
        \midrule
        w/o Strategy    & 64.9 & 22.2 & 43.6\\
        w/o Analogy     & 64.7 & 21.0 & 42.9\\
        w/o Restate     & 61.6 & 22.0 & 41.8\\
        \midrule
        MetaLadder          & \textbf{66.2} & \textbf{22.4} & \textbf{44.3}\\
        \bottomrule
    \end{tabular}
    }
    \caption{Ablation study on GSM8K and MATH, where w/o Strategy, w/o Meta-problem, and w/o Restate refer to the absence of the problem type and solution method, analogy meta-problem, and problem restating mechanism, respectively.}
    \label{exp:ablation}
\end{table}

As shown in Table~\ref{exp:ablation}, we observed that removing any of these components resulted in a significant performance drop across both datasets. 
Specifically, removing the strategy component caused a 0.7\% average decrease in performance on both datasets, indicating that the strategy is important in guiding the model toward more accurate and efficient solutions. Furthermore, excluding the analogy meta-problem or the problem restating mechanism led to even greater performance degradation, with decreases of 1.4 and 2.5 points, respectively. This highlights the crucial role of these components in enhancing the model’s reasoning ability.

\section{Analysis}

\begin{figure*}[h]
    \centering
    \subfigure{
        \includegraphics[width=0.3\linewidth]{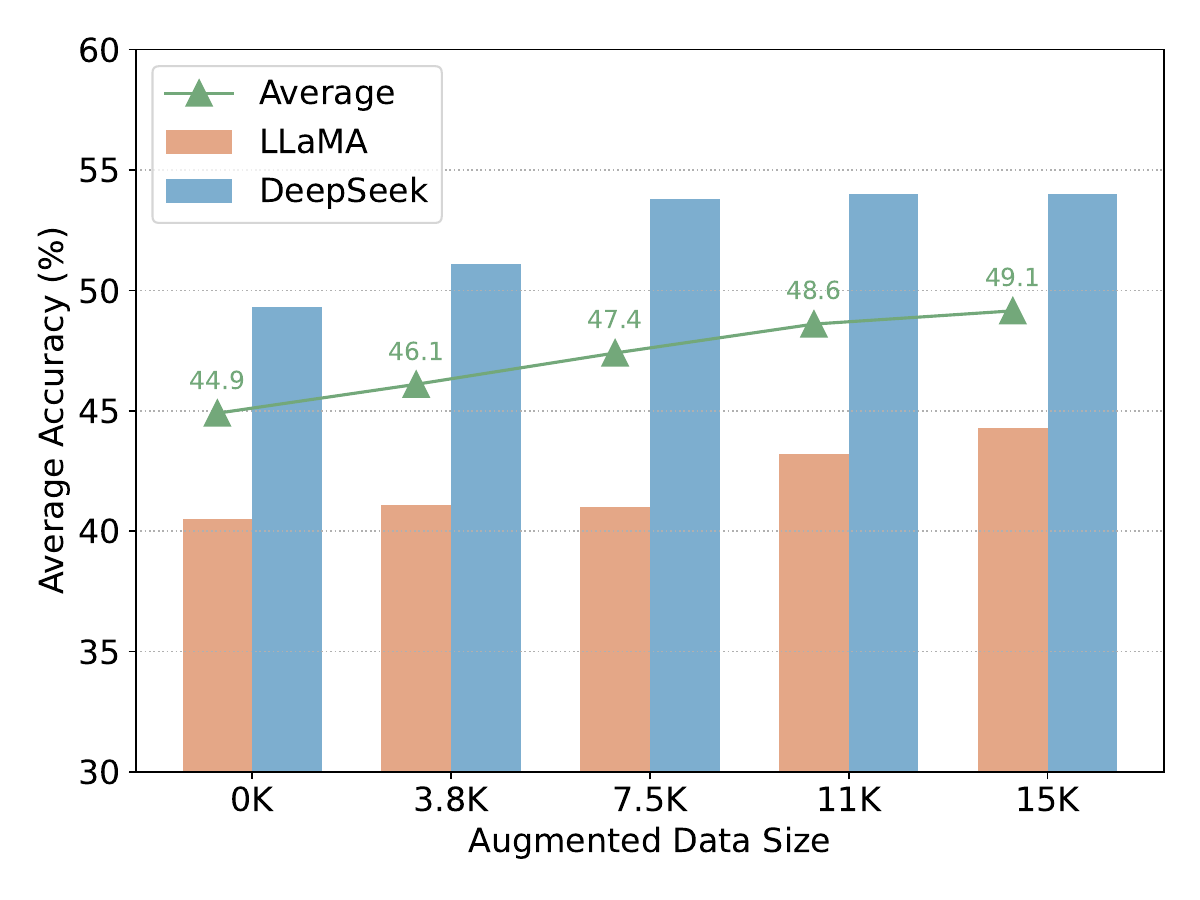}
        \label{fig:amount}
    }
    \subfigure{
        \includegraphics[width=0.3\linewidth]{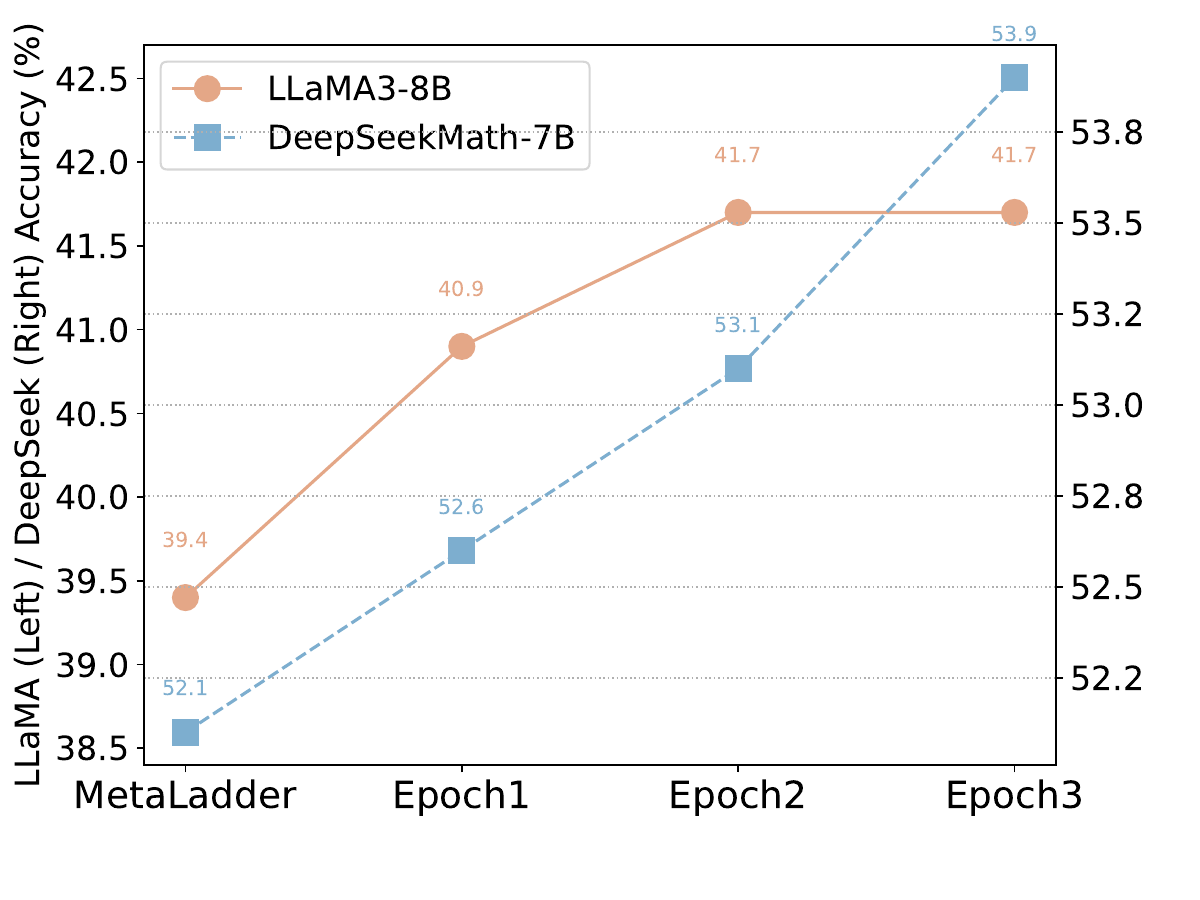}
        \label{fig:evolve}
    }
    \subfigure{
        \includegraphics[width=0.3\linewidth]{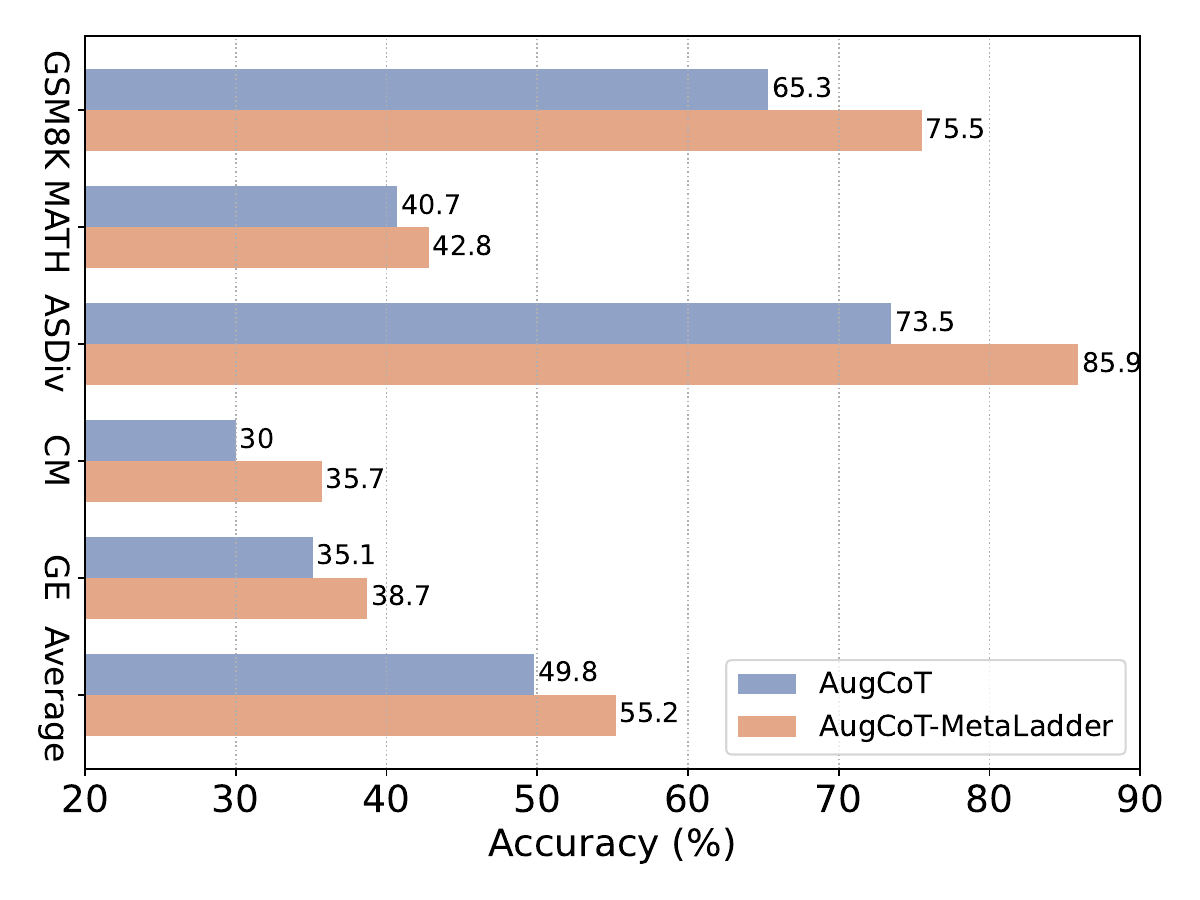}
        \label{fig:combine}
    }
    \caption{\textbf{Left}: Performance of enhancing different amounts of original data. \textbf{Middle}: Results of 3 rounds of evolution on LLaMA and DeepSeek.
\textbf{Right}: Combination of MetaLadder with AugCoT method.}
    \label{fig:all}
\end{figure*}
\subsection{The Amount of MetaLadder Data}

We first investigate the impact of different amounts of MetaLadder-enhanced data on model performance across two models.
As shown in Figure \ref{fig:amount}, in our experiments, we gradually replace the original CoT data with MetaLadder-enhanced data. As the proportion of MetaLadder-augmented data increased, the model's performance steadily improves, reaching its peak when the original data is completely replaced with augmented data. This demonstrates the effectiveness and scalability of the augmentation method.

\subsection{Multi-round Self-evolve}
To further explore the impact of multiple rounds of self-evolution, we examine the performance improvements with each additional round of self-training. As shown in Figure~\ref{fig:evolve}, the performance of MetaLadder steadily improves with the increasing number of self-evolution rounds. On the LLaMA3-8B model, after one round of self-evolution, the average accuracy across all testsets increases from 39.4 to 40.9. After two rounds, the accuracy further improves to 41.7, with an increase of 2.3 points. On the DeepSeekMath-7B model, after two and three rounds of self-evolution, the average accuracy increases to 53.1 and 53.9, respectively, with improvements of 1.0 points and 1.8 points. This demonstrates that multi-round self-evolution significantly enhances the model's reasoning ability. Further experimental results can be found in the Table~\ref{exp:evolve}.

\subsection{Impact of Reflection on Train and Test}

To investigate the impact of reflection on model performance, we compared the effects of performing reflection before and after generating the final answer. We train Pre-RefAug by shifting the reflection component of RefAug to the training stage, and Post-MetaLadder by placing the reflection component of MetaLadder after answer generation. 

As shown in Table~\ref{tab:reflection}, Pre-RefAug outperformed RefAug by 1.4 and 2.1 points on two models, while MetaLadder achieved scores 1.3 and 0.3 points higher than Post-MetaLadder. Our results demonstrate that allowing reflection before providing the final answer leads to further performance improvement. This suggests that reflection data not only enhances the model's reflective capabilities during training but also guides its reasoning during testing in an in-context learning manner, ultimately boosting performance.

\begin{table*}[]
\centering
\resizebox{0.7\textwidth}{!}{
\begin{tabular}{llcccc}
\toprule
  \multicolumn{1}{c}{\multirow{2}{*}{\textbf{Method}}}&
  \multicolumn{1}{c}{\multirow{2}{*}{\textbf{Train $\to$ Inference}}} &
  \multicolumn{2}{c}{\textbf{LLaMA}} &
  \multicolumn{2}{c}{\textbf{DeepSeek}} \\  
  \cmidrule(rl){3-4} \cmidrule(rl){5-6}
 &
  \multicolumn{1}{c}{} &
  \multicolumn{1}{l}{GSM8K} &
  \multicolumn{1}{l}{MATH} &
  \multicolumn{1}{l}{GSM8K} &
  \multicolumn{1}{l}{MATH} \\ \midrule
RefAug          & $QCR$ $\to$ $QC$            & 59.7 & 20.3 & 67.4 & 35.1 \\ \rowcolor{gray!20}
Pre-RefAug      & $QRC$ $\to$ $QRC$       & \textbf{61.0} & \textbf{21.7} & \textbf{69.1} & \textbf{37.5} \\ \midrule
Post-Metaladder & $QQCSQ'C'$ $\to$ $QQC$      & 64.1 & 21.9 & \textbf{70.3} & 37.1 \\ \rowcolor{gray!20}
MetaLadder      & $QSQ'C'QC$ $\to$ $QSQ'C'QC$ & \textbf{66.2} & \textbf{22.4} & 69.4 & \textbf{38.6} \\ \bottomrule
\end{tabular}
}
\caption{Comparison of model performance across different methods with varying placements of reflections before and after the answer.}
\label{tab:reflection}
\end{table*}

\subsection{Combine with Other Augmentation}

We further investigate the effectiveness of our method when combined with other augmentation approaches.
We compare the performance before and after applying the MetaLadder method on two augmented training sets: 
1) In the AugCoT method, we enhance the original solutions using prompts that are almost identical to those used for generating MetaLadder's analogous data, aiming to better align the distribution and complexity of the answers with the meta-problem. 2) We also use data generated by the mainstream MetaMath method~\cite{metamath}, which employs both question rephrasing and answer augmentation to generate a more diverse set of problems.

The results in Figure~\ref{fig:combine} demonstrate the effectiveness of the MetaLadder framework in improving the performance of DeepSeekMath-7B on mathematical benchmark tasks. 
Compared to the AugCoT baseline, AugCoT-MetaLadder shows a more significant improvement across all datasets, with an average increase of 5.4 points in accuracy. This result suggests that the performance boost brought by MetaLadder is not mainly because of the enhanced data only, but rather due to the structured, reflective data generation process within the MetaLadder framework. 

Additionally, as shown in Table~\ref{tab:combine}, we present the experimental results based on the augmented data from MetaMath20K and MetaMath40K. After being enhanced with MetaLadder, MetaMath20K-MetaLadder outperforms the original MetaMath20K with an average performance improvement of 1.3 points, which highlights the positive impact of MetaLadder on model accuracy.

These results further suggest that combining MetaLadder with other augmentation methods can more effectively boost the model's performance, demonstrating the potential of structured data augmentation in improving mathematical reasoning.

\subsection{Case Study}
\begin{table}[h]
\begin{exmp}{Case in Math: Polynomial Roots Problem}{original_questions}
\small
\textbf{Question}: 
Compute \(a + b + c\), given that \(a\), \(b\), and \(c\) are the roots of
\(\frac{1}{x} + 5x^2 = 6x - 24.\)
\vspace{1mm}

\textcolor{red}{\textbf{CoT}:} We can rewrite the equation as $5x^3 - 6x^2 - x + 24 = 0.$
By the rational root theorem, the only possible rational roots are $\pm 1,$ $\pm 2,$ $\pm 3,$ $\pm 4,$ $\pm 6,$ $\pm 8,$ $\pm 12,$ and $\pm 24.$
We can check that $x = 2$ is a root, so $5x^2 - 13x + 12 = 0.$
The roots of this quadratic are 
\(
x = \frac{13 \pm \sqrt{169 - 240}}{10} = \frac{13 \pm \sqrt{-71}}{10} = \frac{13 \pm i \sqrt{71}}{10}.
\)
Thus, the sum of the roots is 
\(2 + \frac{13 + i \sqrt{71}}{10} + \frac{13 - i \sqrt{71}}{10} = 3.\)

The answer is $\textcolor{Red}{\boxed{3}}$.
\vspace{1mm}

\textcolor{OliveGreen}{\textbf{MetaLadder}:} This is a polynomial equation problem where we need to find the sum of the roots of a polynomial. The solution method involves rearranging the equation to a standard form and using Vieta's formulas to find the sum of the roots.
\vspace{1mm}

\textbf{Similar Problem}:
Compute $a+b+c,$ given that $a,$ $b,$ and $c$ are the roots of 
\(\frac{1}{x} + 4x^2 = 5x - 15.\)
\vspace{1mm}

\textbf{Solution to the Similar Problem}:
We can rewrite the equation as 
\(
4x^3 - 5x^2 + 5x - 1 = 0.
\)
By Vieta's formulas, the sum of the roots is $\frac{5}{4}.$
\vspace{1mm}

\textbf{Original Problem}: ...
\vspace{1mm}

\textbf{Solution to the Original Problem}:
We can rewrite the equation as 
\(
5x^3 - 6x^2 + 6x - 1 = 0.
\)
By Vieta's formulas, the sum of the roots is $\frac{6}{5}.$
Thus, the answer is $\textcolor{OliveGreen}{\boxed{\frac{6}{5}}.}$

\end{exmp}
\caption{An example of a model prediction on a polynomial roots problem in MATH dataset, demonstrating the process of solving for the sum of the roots using algebraic manipulation and Vieta's formulas.}
\label{exp:main_case}
\end{table}

To have a more straightforward understanding of the advantage of our MetaLadder, we show some cases and make discussions in this section.

In case \ref{exp:main_case} (in Appendix, with more cases in Appendix~\ref{apx:cases}.), we examine the performance of MetaLadder and the standard CoT approach on a root-finding problem. The problem is: \textit{``Compute \(a + b + c\), given that \(a\), \(b\), and \(c\) are the roots of \(\frac{1}{x} + 5x^2 = 6x - 24.\)''}
CoT solves the problem directly, focusing on converting the equation to \(4x^3 - 5x^2 + 5x - 1 = 0\) and immediately applying Vieta's formulas to obtain \(a + b + c = \frac{5}{4}\). This approach involves minimal abstraction and no explicit reuse of methods from other problems. It delivers a direct result but provides limited insight into the generalization of the solution method.
In contrast, the MetaLadder framework explicitly identifies this task as part of a general class of polynomial-related problems and uses Vieta's formulas as the backbone of the solution. In addition, it builds a reusable methodology, highlights similarities with the original problem \((\frac{1}{x} + 5x^2 = 6x - 24)\), and emphasizes systematic computation techniques. The solution still delivers an accurate result \((a + b + c = \frac{5}{4})\), but it also builds a more structured understanding of the type of problem.

This example highlights how MetaLadder can improve the accuracy and reliability of the model in solving  conceptually rich and broadly applicable problems, further underscoring the value of reflective reasoning in enhancing the model’s overall problem-solving capabilities.

\section{Conclusion}
In this paper, we introduce MetaLadder, a novel framework that enhances the mathematical problem-solving abilities of LLMs. By explicitly prompting the model to reflect on analogical problems and their solutions, MetaLadder enables it to transfer reasoning across similar tasks, mimicking human learning. Additionally, our problem-restating mechanism further enhances the model's reasoning accuracy. Experimental results across multiple mathematical benchmarks demonstrate that MetaLadder significantly improves LLM performance, surpassing both standard Chain-of-Thought (CoT) methods and other state-of-the-art approaches. Our work highlights the importance of integrating analogical reasoning and meta-cognitive strategies into LLMs for complex reasoning tasks. 

\section*{Limitations}
Although the MetaLadder framework has shown promising progress in mathematical problem solving, there are still some limitations worth further exploration and improvement. For instance, the performance of MetaLadder relies on the quality of the analogy problems $Q'$ and their corresponding solutions $C'$. During the generation of analogy problems, data augmentation biases may be introduced, especially when the analogy problems are generated with a strong reliance on certain problem types or solution methods. The model may overfit to common problem types or solutions present in the training data, potentially impacting its ability to generalize to novel problems. Future work could focus on improving the quality of generated analogy problems, enhancing the model's ability to handle a wider variety of problem types, and further investigating the trade-off between inference efficiency and reasoning depth.

\bibliography{custom}

\clearpage
\appendix

\section{Experimental Details}
\subsection{Annotation Details}
\label{apx:gen_detail}
Data annotation was performed using the GPT-4o-mini-2024-07-18 model API with a sampling temperature of 0.7. The full prompt used for annotation is shown in Block \ref{annotation_prompt}.

In Experiment~\ref{tab:combine}, we use prompt shown in Block~\ref{rephrasing_prompt} to rephrase the original solutions in GSM8K and MATH.

\subsection{Train Details}
\label{apx:train_detail}
Model training was conducted using the LLaMA Factory~\footnote{https://github.com/hiyouga/LLaMA-Factory} on 8 NVIDIA A100 GPUs. 
We train all models for one epoch with a batch size of 128, using the AdamW optimizer~\cite{adamw} with a learning rate of 5e-6 and cosine learning rate decay.
The training prompt is shown in Block \ref{train_prompt}.

\subsection{Evaluation Details}

\noindent\textbf{GSM8K~\cite{cobbe2021gsm8k}}:
GSM8K consists of grade-school arithmetic tasks with relatively low difficulty, primarily used to evaluate basic mathematical reasoning abilities. The testset includes 1319 basic math word problems, covering simple arithmetic operations such as addition, subtraction, multiplication, and division. Compared to other more complex datasets, the problems in GSM8K are straightforward and suitable for testing a model's performance on solving basic mathematical problems.

\noindent\textbf{MATH~\cite{hendrycks2021math}}:
The MATH testset contains 5000 challenging competition-level math problems. These problems are designed to be complex and require the model to possess higher-level mathematical reasoning capabilities, far surpassing the simpler problems found in GSM8K. MATH spans multiple mathematical domains, including algebra, geometry, and number theory, making it an ideal benchmark for evaluating a model’s performance on complex mathematical reasoning tasks.

\noindent\textbf{ASDiv~\cite{miao2020diverse}}:
ASDiv (Academia Sinica Diverse MWP Dataset) is a diverse English math word problem dataset intended to evaluate the capabilities of various MWP solvers. This dataset includes 2,305 math word problems that cover a wide range of language patterns and problem types, offering more diversity than existing MWP datasets. It includes problems commonly found in elementary school and is annotated with problem types and grade levels to help assess the difficulty and complexity of each problem.

\noindent\textbf{Gaokao 2023 EN~\cite{liao2024mario}}:
Gaokao 2023 EN contains 385 math problems from the 2023 Chinese National College Entrance Examination (Gaokao), which are primarily high school-level open-ended problems. These problems cover a wide range of mathematical topics and include content taught during high school in China. The Gaokao EN2023 dataset is designed to assess students’ ability to apply mathematical reasoning in real-world situations, containing both basic problems and more complex ones. It serves as an important benchmark for evaluating models' performance on Gaokao-style math problems.

\noindent\textbf{CollegeMath~\cite{tang2024mathscale}}:
The CollegeMath dataset includes 2,818 college-level math problems extracted from 9 textbooks, spanning 7 mathematical domains such as linear algebra and differential equations. CollegeMath is designed to test a model’s ability to reason across diverse mathematical topics, with a particular focus on generalization to complex mathematical reasoning tasks at the college level. The problems are more difficult, making the dataset well-suited for evaluating a model's ability to solve advanced mathematical problems.

\noindent\textbf{DeepMind-Mathematics~\cite{mathematic}}: The DeepMind-Mathematics test set containing 1000 problems from a variety of problem types, based on a national school mathematics curriculum (up to age 16), designed to assess basic mathematical reasoning across different domains. The dataset generates question and answer pairs of varying types, generally at school-level difficulty, and aims to test the mathematical learning and algebraic reasoning abilities of learning models.

\label{apx:evaluation_detail}
All model evaluation was carried out using the framework provided at \href{https://github.com/ZubinGou/math-evaluation-harness/tree/main}{https://github.com/ZubinGou/math-evaluation-harness/tree/main} with zero-shot evaluation, greedy sampling and a maximum generation length of 2048 tokens.

To validate that the improvement from the shortcut method was not due to avoiding the truncation of MetaLadder's output, we also conducted the main experiment with a maximum length of 4096 tokens, and no significant changes in the metrics were observed.

\section{More Experiments}

\begin{table*}[]
\centering
\renewcommand{\arraystretch}{0.9}
\small
\resizebox{\textwidth}{!}{
\begin{tabular}{lcccccccc}
\toprule
\multicolumn{2}{c}{}  & \multicolumn{2}{c}{\textbf{In-Domain}} & \multicolumn{4}{c}{\textbf{Out-of-Domain}}  \\ \cmidrule(rl){3-4} \cmidrule(rl){5-8} 
\multicolumn{1}{c}{\multirow{-2}{*}{\textbf{Method}}} & \multicolumn{1}{c}{\multirow{-2}{*}{\textbf{\# Sample}}} & GSM8K     & MATH     & ASDiv & CM & GE & DM & Average\\ \hline
\multicolumn{9}{c}{\cellcolor[HTML]{C0C0C0}\textbf{\textit{LLaMA3-8B}}}       \\
\hline
MetaLadder                &15K  & 66.2     & 22.4    & 76.7  & 17.2 & 23.9 & 29.7 & 39.4\\
MetaLadder+Self-evolve    &22K  & 66.5     & 24.8    & 76.6  & 19.1 & 26.8 & 31.7 & 40.9\\
MetaLadder+Self-evolve2 &30K &68.9 &25.1 &79.3 &19.1 &25.7 &32.2 &41.7\\
MetaLadder+Self-evolve3 &38K &69.2 &25.9 &77.9 &19.8 &26.0 &31.6 &41.7\\
\hline
\multicolumn{9}{c}{\cellcolor[HTML]{C0C0C0}\textbf{\textit{DeepSeekMath-7B}}} \\
\hline
MetaLadder                   &15K  & 69.4  & 38.6  & 85.9 & 32.6 & 37.4 & 48.4 & 52.1\\
MetaLadder+Self-evolve         &23K  & 70.5 & 39.3 & 86.3 & 33.3 & 35.1 & 50.8 & 52.6\\
MetaLadder+Self-evolve2        &31K & 71.9 & 39.8 & 86.1 & 33.6 & 37.9 & 49.2 & 53.1\\
MetaLadder+Self-evolve3        &40K & 72.2 & 40.6 & 86.3 & 34.0 & 38.7 & 51.5 & 53.9\\
\bottomrule
\end{tabular}
}
\caption{\textbf{Accuracy of self-evolution on in-domain and out-of-domain mathematical benchmarks}. The \textbf{bold} and \underline{underlined} values represent the first and second best performances, respectively. CM, GE, DM denotes College Math, Gaokao En 2023, DeepMind-Mathematics, respectively.}
\label{exp:evolve}
\end{table*}

\begin{table*}[]
\centering
\renewcommand{\arraystretch}{0.9}
\begin{tabular}{lccccccc}
\toprule
\textbf{Method} & \textbf{GSM8K} & \textbf{MATH} & \textbf{ASDiv} & \textbf{CM} & \textbf{GE} & \textbf{DM} & \textbf{Average} \\ \midrule
MetaMath20K            & 71.1 & 38.4 & 84.5 & \underline{32.0} & 32.5 & 46.9 & 50.9 \\
MetaMath20K-MetaLadder & 72.9 & 39.8 & \underline{86.5} & 31.6 & 34.3 & 44.9 & 51.7 \\ \hdashline
MetaMath40K            & \underline{73.9} & 38.5 & 84.6 & \underline{32.0} & 34.3 & 48.2 & 51.9 \\
MetaMath40K-MetaLadder & \textbf{75.7} & 40.0 & \textbf{87.0} & 30.5 & 34.8 & 45.8 & \underline{52.3} \\ \hdashline
AugCoT                 & 65.3 & \underline{40.7} & 73.5 & 30.0 & \underline{35.1} & \textbf{54.0} & 49.8  \\
AugCoT-MetaLadder      & 75.5 & \textbf{42.8} & 85.9 & \textbf{35.7} & \textbf{38.7} & \underline{52.8} & \textbf{55.2}\\ 
\bottomrule
\end{tabular}
\caption{\textbf{DeepSeekMath-7B performance on two augmented datasets}. MetaMath20K constructed by uniformly sampling half of the data from MetaMath40K, and AugCoT representing original solutions rephrased by 4o-mini to match the style of analogous data used in MetaLadder.}
\label{tab:combine}
\end{table*}

\begin{table*}[h]
    \centering
    \begin{tabular}{lccccc}
        \toprule
        \textbf{Method} & \textbf{Cos} & \textbf{JD} & \textbf{LD} & \textbf{GSM8K} & \textbf{MATH}\\
        \midrule
        Original Problem & 1.00 & 0.00 & 0.00  & 64.0 & 21.1 \\
        Analogous Problem & 0.93 & 0.48 & 101.00 & 66.2  & 22.4 \\
        Enhanced Analogous Problem & 0.91 & 0.61 & 131.30 & 65.3  & 23.3 \\
        \bottomrule
    \end{tabular}
    \caption{The impact of repeating the original problem and using analogous problems with greater divergence on model performance. Cos refers to Cosine Similarity, JD refers to Jaccard Distance, and LD refers to Levenshtein Distance.}
    \label{exp:sim}
\end{table*}

\begin{table}[h]
    \centering
    \begin{tabular}{lrr}
        \toprule
        \textbf{Method} & \textbf{GSM8K} & \textbf{MATH} \\
        \midrule
        LLaMA-CoT                   & 61.86 & 749.26\\
        LLaMA-MetaLadder            & 172.62 & 1479.85\\
        LLaMA-MetaLadder-cut       & 111.26 & 1081.58\\
        \midrule
        DeepSeek-CoT                & 65.9 & 1253.26\\
        DeepSeek-MetaLadder         & 210.15 & 2181.13\\
        DeepSeek-MetaLadder-cut    & 113.96 & 1343.74\\
        \bottomrule
    \end{tabular}
    \caption{Total time cost of inference on the whole GSM8K and MATH testsets, in seconds. Our MetaLadder, combined with shortcut reasoning, significantly reduces inference time, achieving speeds close to CoT.}
    \label{exp:time}
\end{table}

\clearpage
\begin{table*}[h]
\begin{tcolorbox}[colback=gray!5,colframe=black!75, width=\textwidth, title=Data Annotation Prompt]
\small
You are a professional math teacher, and your goal is to teach your student to learn a given math problem. Identify the type of the given problem and the name of the solution method. Then, generate a similar problem with its solution.
\vspace{2mm}

\#\# Example 1: \\
\#\#\# Original Problem: \\
Youngsville had a population of 684 people. The town had a growth spurt and the population increased by 25\% then they witnessed that 40\% of the population moved away. What is the current population?
\vspace{2mm}

\#\#\# Solution to the Original Problem: \\
The town had 684 people, and then had a 25\% growth spurt, so the population increased by 684 * 0.25 = 171 people. This increase brought the population to 684 + 171 = 855 people. 40\% of the population moved away, so 855 * 0.40 = 342 people moved away. The new population is 855 - 342 = 513 people. The answer is 513.
\vspace{2mm}

\#\#\# Type of Problem and Solution Method: \\
This is a consecutive percentage change problem. The solution method involves applying the growth factor first, then applying the reduction factor. The key to solve the problem is to understand the concept of relative increase and decrease percentages.
\vspace{2mm}

\#\#\# Similar Problem: \\
A company started with 800 employees. After one year, the workforce increased by 20\%, and then 30\% of the new workforce decided to quit due to relocation. How many employees remain at the company?
\vspace{2mm}

\#\#\# Solution to the Similar Problem: \\
1. Initial employees: 800.

2. Increase by 20\%: 20\% of 800 is 160, so the workforce grows to 800 + 160 = 960.

3. Then, 30\% of these 960 employees quit: 30\% of 960 is 288, so 288 employees leave.

4. Remaining employees: 960 - 288 = 672. Therefore, the final number of employees is 672.

\vspace{1mm}

\#\# Example 2: \\
\#\#\# Original Problem: \\
Solve the equation (x - 99)(x - 101) = 8.
\vspace{2mm}

\#\#\# Solution to the Original Problem: \\
Let t=x-100. Then the equation becomes (t - 1)(t + 1) = 8, which transforms into $t^2$ - 1 = 8. Therefore, t = 3 or t = -3, and accordingly we get x = 97 or x = 103.
\vspace{2mm}

\#\#\# Type of Problem and Solution Method: \\
This is a quadratic equation problem solved by introducing a substitution to simplify the expression. The solution method involves recognizing a suitable substitution that transforms the equation into a simpler form. The key to solving the problem is understanding how to use algebraic manipulation to facilitate solving equations.

\vspace{2mm}

\#\#\# Similar Problem: \\
Solve the equation (x - 50)(x - 52) = 4.
\vspace{2mm}

\#\#\# Solution to the Similar Problem: \\
1. Notice that the middle point between 50 and 52 is 51, so let t = x - 51.

2. Then (x - 50) = (t + 1) and (x - 52) = (t - 1).

3. The equation becomes (t + 1)(t - 1) = 4 $\Rightarrow$ $t^2$ - 1 = 4 $\Rightarrow$ $t^2$ = 5 $\Rightarrow$ t = $\sqrt{5}$ or t = $-\sqrt{5}$.

4. Substituting back: x = t + 51. \\
    - If $t = \sqrt{{5}}$, then $x = 51 + \sqrt{{5}}$. \\
    - If $t = -\sqrt{{5}}$, then $x = 51 - \sqrt{{5}}$.
   
Hence, the solutions are $x = 51 + \sqrt{5}$ or $x = 51 - \sqrt{5}$.
\vspace{2mm}

\#\#\# Original Problem:

\{problem\}

\vspace{2mm}
\#\#\# Solution to the Original Problem:

\{solution\}

\end{tcolorbox}
\caption{Data annotation prompt for generating and solving similar math problems. The prompt provides a structured approach for identifying problem types, solution methods, and creating analogous problems with detailed solutions.}
\label{annotation_prompt}
\end{table*}

\begin{table*}[h]
\begin{tcolorbox}[colback=gray!5,colframe=black!75, width=\textwidth, title=Answer Augmentation Prompt]
\small
You are a professional math teacher, and your goal is to teach your student to learn a given math problem. Identify the type of the given problem and the name of the solution method. Then, generate a similar problem with its solution.
\vspace{2mm}

\#\# Example 1: \\
\#\#\# Problem:
Youngsville had a population of 684 people. The town had a growth spurt and the population increased by 25\% then they witnessed that 40\% of the population moved away. What is the current population?
\vspace{2mm}

\#\#\# Original Solution:\\
The town had 684 people, and then had a 25\% growth spurt, so the population increased by 684 * 0.25 = 171 people. This increase brought the population to 684 + 171 = 855 people. 40\% of the population moved away, so 855 * 0.40 = 342 people moved away. The new population is 855 - 342 = 513 people. The answer is 513.
\vspace{2mm}

\#\#\# Augmented Solution:\\
1. Initial population: 684.\\
2. Growth by 25\%: 25\% of 684 is 171, so the population increases to 684 + 171 = 855.\\
3. 40\% of 855 moved away: 40\% of 855 is 342, so 342 people moved away.\\
4. The remaining population is 855 - 342 = 513.\\ 
Therefore, the final population is 513.
\vspace{2mm}

\#\# Example 2: \\
\#\#\# Problem: \\
Solve the equation (x - 99)(x - 101) = 8.
\vspace{2mm}

\#\#\# Original Solution: \\
Let $t = x - 100$. Then the equation becomes $(t - 1)(t + 1) = 8$, which transforms into $t^2 - 1 = 8$. Therefore, $t = 3$ or $t = -3$, and accordingly we get x = 97 or x = 103.\\

\#\#\# Augmented Solution: \\
1. Notice that the middle point between 99 and 101 is 100, so let t = x - 100.

2. Then $(x - 99) = (t + 1)$ and $(x - 101) = (t - 1)$.

3. The equation becomes $(t + 1)(t - 1) = 8 \Rightarrow t^2 - 1 = 8 \Rightarrow t^2 = 9 \Rightarrow t = 3 or t = -3$.

4. Substituting back: x = t + 100.\\
   - If t = 3, then x = 100 + 3.\\
   - If t = -3, then x = 100 - 3.
   
Hence, the solutions are x = 103 or x = 97.
\vspace{2mm}

\#\#\# Problem:

\{problem\}\\

\#\#\# Original Solution:

\{solution\}

\end{tcolorbox}
\caption{Answer augmentation prompt designed for creating detailed step-by-step solutions to math problems.In the AugCoT method, we enhance the original solutions using this prompt, which is almost identical to those used for generating MetaLadder's analogous data, aiming to better align the distribution and complexity of the answers with the meta-problem.}
\label{rephrasing_prompt}
\end{table*}

\begin{table*}[h]
\begin{tcolorbox}[colback=gray!5,colframe=black!75, width=\textwidth, title=Training and Evaluation Prompt]
\small
\#\#\# Instruction: \\
\{question\}
\vspace{2mm}

\#\#\# Response: Let's think step by step.

\end{tcolorbox}
\caption{Training and evaluation prompt designed to guide the model in generating step-by-step responses to math problems.}
\label{train_prompt}
\end{table*}

\clearpage
\section{More cases}
We present more cases in this section to show the generated predictions of our MetaLadder trained model. 
\label{apx:cases}
\begin{table*}[h]
\label{apx:case1}
\vspace{2mm}
\begin{exmp}{Case in MATH}{case1}
\small
\textbf{Question}: Exits on a highway are numbered consecutively from 1 to 50. The distance from exit 41 to exit 50 is 100 km. If each exit is at least 6 km from the next exit, what is the longest possible distance, in kilometers, between exit 47 and exit 48?\\\\
\textcolor{red}{\textbf{CoT}:} The distance between exit 41 and exit 50 is 100 km. If each exit is at least 6 km from the next exit, the distance between exit 47 and exit 48 is at least 12 km. The longest possible distance between exit 47 and exit 48 is 12 km.\\
Final Answer: \textcolor{red}{\boxed{12}.} \\\\
\textcolor{OliveGreen}{\textbf{MetaLadder}:} This is a problem involving the arrangement of objects (exits) with constraints on the distances between them. The solution method involves setting up inequalities based on the given conditions and then solving for the maximum distance between two specific exits.\\

Similar Problem:

Exits on a highway are numbered consecutively from 1 to 40. The distance from exit 31 to exit 40 is 80 km. If each exit is at least 5 km from the next exit, what is the longest possible distance, in kilometers, between exit 37 and exit 38?\\

Solution to the Similar Problem:

1. The distance from exit 31 to exit 40 is 80 km, which means the distance from exit 31 to exit 32 is at least 5 km.

2. The distance from exit 32 to exit 33 is at least 5 km, and so on.

3. The total distance from exit 31 to exit 38 is:
\begin{itemize}
\setlength{\topsep}{0pt}  
\setlength{\itemsep}{0pt}
\setlength{\parsep}{0pt}
\setlength{\parskip}{0pt}
    \item From exit 31 to exit 32: 5 km
    \item From exit 32 to exit 33: 5 km
    \item From exit 33 to exit 34: 5 km
    \item From exit 34 to exit 35: 5 km
    \item From exit 35 to exit 36: 5 km
    \item From exit 36 to exit 37: 5 km
    \item From exit 37 to exit 38: 5 km
\end{itemize}
4. Therefore, the total distance from exit 31 to exit 38 is:

\[
5 \text{ km} \times 7 = 35 \text{ km}
\]

5. The distance from exit 38 to exit 40 is:

\[
80 \text{ km} - 35 \text{ km} = 45 \text{ km}
\]
6. The distance from exit 37 to exit 38 is:

\[
45 \text{ km} - 5 \text{ km} = 40 \text{ km}
\]

Thus, the longest possible distance between exit 37 and exit 38 is **40 km**.\\

Original Problem:

Exits on a highway are numbered consecutively from 1 to 50. The distance from exit 41 to exit 50 is 100 km. If each exit is at least 6 km from the next exit, what is the longest possible distance, in kilometers, between exit 47 and exit 48?\\

Solution to the Original Problem:

The distance from exit 41 to exit 50 is 100 km, which means the distance from exit 41 to exit 42 is at least 6 km.

The distance from exit 42 to exit 43 is at least 6 km, and so on.

The total distance from exit 41 to exit 48 is:
\begin{itemize}
\setlength{\topsep}{0pt}  
\setlength{\itemsep}{0pt}
\setlength{\parsep}{0pt}
\setlength{\parskip}{0pt}
\item From exit 41 to exit 42: 6 km
\item From exit 42 to exit 43: 6 km
\item From exit 43 to exit 44: 6 km
\item From exit 44 to exit 45: 6 km
\item From exit 45 to exit 46: 6 km
\item From exit 46 to exit 47: 6 km
\item From exit 47 to exit 48: 6 km
\end{itemize}
Therefore, the total distance from exit 41 to exit 48 is:

6 km * 7 = 42 km

The distance from exit 48 to exit 50 is:

100 km - 42 km = 58 km

The distance from exit 47 to exit 48 is:

58 km - 6 km = 52 km

Final Answer: \textcolor{OliveGreen}{\boxed{52}}.

\end{exmp}
\vspace{2.5mm}

\end{table*}

\begin{table*}[h]
\vspace{2mm}
\begin{exmp}{Case in MATH: Trigonometry}{case2}
\small
\textbf{Question}: Find $\csc (-120^\circ)$.\\\\
\textcolor{red}{\textbf{CoT}:} We have $\csc (-120^\circ) = \frac{1}{\sin (-120^\circ)} = \frac{1}{\sin (120^\circ)} = \frac{1}{\frac{\sqrt{3}}{2}} = \boxed{\frac{2\sqrt{3}}{3}}$.\\
Final Answer: \textcolor{red}{\boxed{\frac{2\sqrt{3}}{3}}}.\\\\
\textcolor{OliveGreen}{\textbf{MetaLadder}:} This is a trigonometric function evaluation problem. The solution method involves using the properties of the cosecant function and the reference angle to find the value of the cosecant at a given angle.\\

Similar Problem:

Find $\csc (-135^\circ)$.\\

Solution to the Similar Problem:

1. The angle $-135^\circ$ is in the third quadrant, where the cosecant function is negative. 

2. The reference angle for $-135^\circ$ is $135^\circ - 180^\circ = -45^\circ$.  

3. The cosecant of $-45^\circ$ is $\csc (-45^\circ) = -\sqrt{2}$.  

4. Therefore, $\csc (-135^\circ) = -\sqrt{2}$.

Thus, the final answer is $ \boxed{-\sqrt{2}}$.\\

Original Problem:

Find $\csc (-120^\circ)$.\\

Solution to the Original Problem:

The angle $-120^\circ$ is in the third quadrant, where the cosecant function is negative. The reference angle for $-120^\circ$ is $120^\circ - 180^\circ = -60^\circ$. The cosecant of $-60^\circ$ is $\csc (-60^\circ) = -\frac{2}{\sqrt{3}}$. Therefore, $\csc (-120^\circ) = \boxed{-\frac{2}{\sqrt{3}}}$.

Final Answer: \textcolor{OliveGreen}{\boxed{-\frac{2}{\sqrt{3}}}}.
\end{exmp}
\vspace{2.5mm}
\end{table*}

\begin{table*}[h]
\vspace{2mm}
\begin{exmp}{Case in GSM8K: Algebra}{case3}
\small
\textbf{Question}: A football team played 22 games. They won 8 more than they lost. How many did they win?\\\\
\textcolor{red}{\textbf{CoT}:} If they won 8 more than they lost, then they lost 22-8=14 games.  
So they won 22-14=8 games.  

Final Answer: \textcolor{red}{\boxed{8}}.\\\\
\textcolor{OliveGreen}{\textbf{MetaLadder}:} This is a problem involving a system of equations based on relationships between wins, losses, and total games played. The solution method involves setting up equations based on the given information and solving for the unknowns.\\

Similar Problem:

A basketball team played 30 games. They won 10 more than they lost. How many did they win?\\

Solution to the Similar Problem:

1. Let the number of games won be \( w \).  

2. The number of games lost is \( w - 10 \).  

3. The total number of games played is 30, so we have the equation:  
   \[
   w + (w - 10) = 30
   \]
4. Simplifying this gives:  
   \[
   2w - 10 = 30
   \]
5. Adding 10 to both sides:  
   \[
   2w = 40
   \]
6. Dividing by 2:  
   \[
   w = 20
   \]

Thus, the team won 20 games.\\

Original Problem:

A football team played 22 games. They won 8 more than they lost. How many did they win?\\

Solution to the Original Problem:

Let the number of games won be \( w \).  
The number of games lost is \( w - 8 \).  
The total number of games played is 22, so we have the equation:  
\[
w + (w - 8) = 22
\]
\[
2w - 8 = 22
\]
\[
2w = 30
\]
\[
w = 15
\]

Final Answer: \textcolor{OliveGreen}{\boxed{15}}.
\end{exmp}
\vspace{2.5mm}
\end{table*}

\begin{table*}[h]
\vspace{2mm}
\begin{exmp}{Case in Math}{case4}
\small
\textbf{Question}: Simplify \([\dfrac{1}{1 - \tan^2 x} + \dfrac{1}{1 - \cot^2x}]\).\\\\
\textcolor{red}{\textbf{CoT}:} We have\([\dfrac{1}{1 - \tan^2 x} + \dfrac{1}{1 - \cot^2 x}] = \dfrac{1}{\cos^2 x - \sin^2 x} + \dfrac{1}{\sin^2 x - \cos^2 x} = \dfrac{1}{\cos^2 x - \sin^2 x} - \dfrac{1}{\\cos^2 x - \\sin^2 x} = 0\\]\). Final Answer:  \textcolor{red}{\boxed{0}}\\\\
\textcolor{OliveGreen}{\textbf{MetaLadder}:} This is a trigonometric simplification problem. The solution method involves using the Pythagorean identities to express the given expression in terms of sine and cosine, and then simplifying the resulting expression.\\\\

Similar Problem:

Simplify \([\dfrac{1}{1 - \sin^2 x} + \dfrac{1}{1 - \cos^2 x}]\).\\\\

Solution to the Similar Problem:

Using the Pythagorean identity, we have \(1 - \sin^2 x = \cos^2 x \quad \text{and} \quad 1 - \cos^2 x = \sin^2 x\). Thus,\(\dfrac{1}{1 - \sin^2 x} + \dfrac{1}{1 - \cos^2 x} = \dfrac{1}{\cos^2 x} + \dfrac{1}{\sin^2 x} = \dfrac{\sin^2 x + \cos^2 x}{\sin^2 x \cos^2 x} = \dfrac{1}{\sin^2 x \cos^2 x}\).\\\\

Original Problem: 

Simplify \(\left[ \dfrac{1}{1 - \tan^2 x} + \dfrac{1}{1 - \cot^2 x} \right]\).\\\\

Solution to the Original Problem

1. Start with the given expression: \([\dfrac{1}{1 - \tan^2 x} + \dfrac{1}{1 - \cot^2 x}]\). \\
2. Use the Pythagorean identity: \([1 - \tan^2 x = \dfrac{1}{\cos^2 x} \quad \text{and} \quad 1 - \cot^2 x = \dfrac{1}{\sin^2 x}]\).\\
3. Substitute these identities into the expression:  \(\dfrac{1}{\dfrac{1}{\cos^2 x}} + \dfrac{1}{\dfrac{1}{\sin^2 x}} = \cos^2 x + \sin^2 x\).\\
4. Since \(\cos^2 x + \sin^2 x = 1\), we have:   \(\cos^2 x + \sin^2 x = 1.\) \\
5.Final Answer: \textcolor{OliveGreen}{\boxed{1}} \\

\end{exmp}
\vspace{2.5mm}
\label{exp:future_age_case}
\end{table*}

\begin{table}[h]
\vspace{2mm}
\begin{exmp}{Case in MATH: Exponentiation}{case5}
\small
\textbf{Question}: Compute $(2^3)(5^5)$ in your head.\\\\
\textcolor{red}{\textbf{CoT}:} We have $(2^3)(5^5) = (2^3)(5^3)(5^2) = (10^3)(25) = 100000$.  

Final Answer: \textcolor{red}{\boxed{100000}}.\\\\
\textcolor{OliveGreen}{\textbf{MetaLadder}:} This is a multiplication problem involving powers of different bases. The solution method involves simplifying the expression by breaking it down into smaller parts and then multiplying the results.\\

Similar Problem:

Compute $(3^2)(4^3)$ in your head.\\

Solution to the Similar Problem:

1. First, calculate \(3^2\):  
   \[
   3^2 = 9
   \]
2. Next, calculate \(4^3\):  
   \[
   4^3 = 4 \times 4 \times 4 = 16 \times 4 = 64
   \]
3. Now, multiply the results:  
   \[
   9 \times 64 = 576
   \]
Thus, the final answer is \( \boxed{576} \).\\

Original Problem:

Compute $(2^3)(5^5)$ in your head.\\

Solution to the Original Problem:

We have \(2^3 = 8\) and \(5^5 = 5 \cdot 25 \cdot 25 = 5 \cdot 625 = 3125\).  
Thus, the answer is \(8 \cdot 3125 = \boxed{25000}\).

Final Answer: \textcolor{OliveGreen}{\boxed{25000}}.
\end{exmp}
\vspace{2.5mm}
\label{exp:case}
\end{table}

\begin{table}[h]
\vspace{2mm}
\begin{exmp}{Case in GSM8K: Basic Algebra}{case6}
\small
\textbf{Question}: A treasure hunter found a buried treasure chest filled with gems. There were 175 diamonds, 35 fewer rubies than diamonds, and twice the number of emeralds than the rubies. How many of the gems were there in the chest?\\\\
\textcolor{red}{\textbf{CoT}:} 
There were \(175 - 35 = 140\) rubies. \\ 
There were \(140 \times 2 = 280\) emeralds. \\
There were \(175 + 140 + 280 = 600\) gems in the chest. \\ 
Final Answer: \textcolor{red}{\boxed{600}}.\\\\
\textcolor{OliveGreen}{\textbf{MetaLadder}:} This is a problem involving basic arithmetic operations (addition and multiplication) to find the total number of items based on given relationships. The solution method involves calculating the quantities of each type of gem based on the relationships provided.\\

Similar Problem:

A collector found a collection of antique coins. There were 200 silver coins, 40 fewer gold coins than silver coins, and three times the number of copper coins than the gold coins. How many coins were there in total? \\

Solution to the Similar Problem:

1. Silver coins: 200.  \\
2. Gold coins: \(200 - 40 = 160\). \\ 
3. Copper coins: \(160 \times 3 = 480\). \\ 
4. Total coins: \(200 + 160 + 480 = 840\). \\ 
Therefore, the total number of coins in the collection is \( \boxed{840} \).\\

Original Problem:

A treasure hunter found a buried treasure chest filled with gems. There were 175 diamonds, 35 fewer rubies than diamonds, and twice the number of emeralds than the rubies. How many of the gems were there in the chest? \\

Solution to the Original Problem:

There were \(175 - 35 = 140\) rubies. \\
There were \(140 \times 2 = 280\) emeralds. \\  
There were \(175 + 140 + 280 = 595\) gems in the chest. \\ 
Final Answer: \textcolor{OliveGreen}{\boxed{595}}.
\end{exmp}
\vspace{2.5mm}
\label{exp:gem_case}
\end{table}

\end{document}